\documentclass[11pt]{article}
\usepackage[preprint]{acl}   
\usepackage{times}
\usepackage{latexsym}
\usepackage[T1]{fontenc}
\usepackage[utf8]{inputenc}
\usepackage{inconsolata}
\usepackage{graphicx}
\usepackage{booktabs}
\usepackage{pifont}
\usepackage{xcolor}

\newcommand{\rev}[1]{#1}
\definecolor{evgreen}{RGB}{0,120,60}
\newcommand{\ev}[1]{#1}

\newcommand{\bench}{DynaPix}

\newcommand{\nretqueries}{2{,}002}
\newcommand{\ndbimages}{10{,}000}
\newcommand{\blfootnote}[1]{%
  \begingroup
  \renewcommand\thefootnote{}\footnote{#1}%
  \addtocounter{footnote}{-1}%
  \endgroup
}
\title{\bench{}: Can Vision-Language Models Identify the Exact Future?}

\author{Thong Nguyen$^{\dagger\,}$ \quad
Vinh-Hien Do \quad
Quynh Vo \quad
Cong-Duy Nguyen \quad 
See-Kiong Ng \quad \\\\
Centre for AI Research, VinUniversity \\
National University of Singapore
\\\\
Email: \texttt{thong.nguyen@u.nus.edu}
}

\begin{document}
\maketitle
\begin{abstract}
Acting in a physical scene requires knowing its real later state, not a plausible one. Current evaluations often accept words or a realistic-looking image, so the predicted state is never checked against the true one. We introduce \bench{} (\textbf{Dyna}mic \textbf{Pix}els), a benchmark that makes prediction checkable. Given a video clip that stops before a key event and a question about a later moment, a model must pick the true future image from close candidates or a large gallery. The scenes come from a physics simulator, so the correct image and its time are known exactly and the wrong options are deliberately similar. Models often succeed when a visible event marks the target moment, but are near chance when only elapsed time marks it. Gallery search is harder still, as the true image rarely ranks first. People handle the elapsed-time items well, so the difficulty lies with the models, not the questions. Training on scene accounts drawn from the simulator's true record, not a teacher's guess, repairs much of this but not the longer elapsed-time case. \bench{} thus exposes a temporal-anchoring gap: models attach a prediction to an event far better than to time itself.
\end{abstract}
\blfootnote{$^{\dagger}$Corresponding author.}
\section{Introduction}

Concrete physical anticipation is a reliability requirement for vision-language
models (VLMs) used as the perception-and-reasoning core of embodied agents. An
agent facing an unstable
stack cannot act on \textit{``the stack may collapse''}: it needs to know which
objects move and where they come to rest, since a vague forecast can turn a safe
route into a collision. We therefore target predictions that commit to the
specific resulting state of the observed scene, not a plausible description of
what might happen.

\begin{figure*}[t]
\centering
\includegraphics[width=\textwidth]{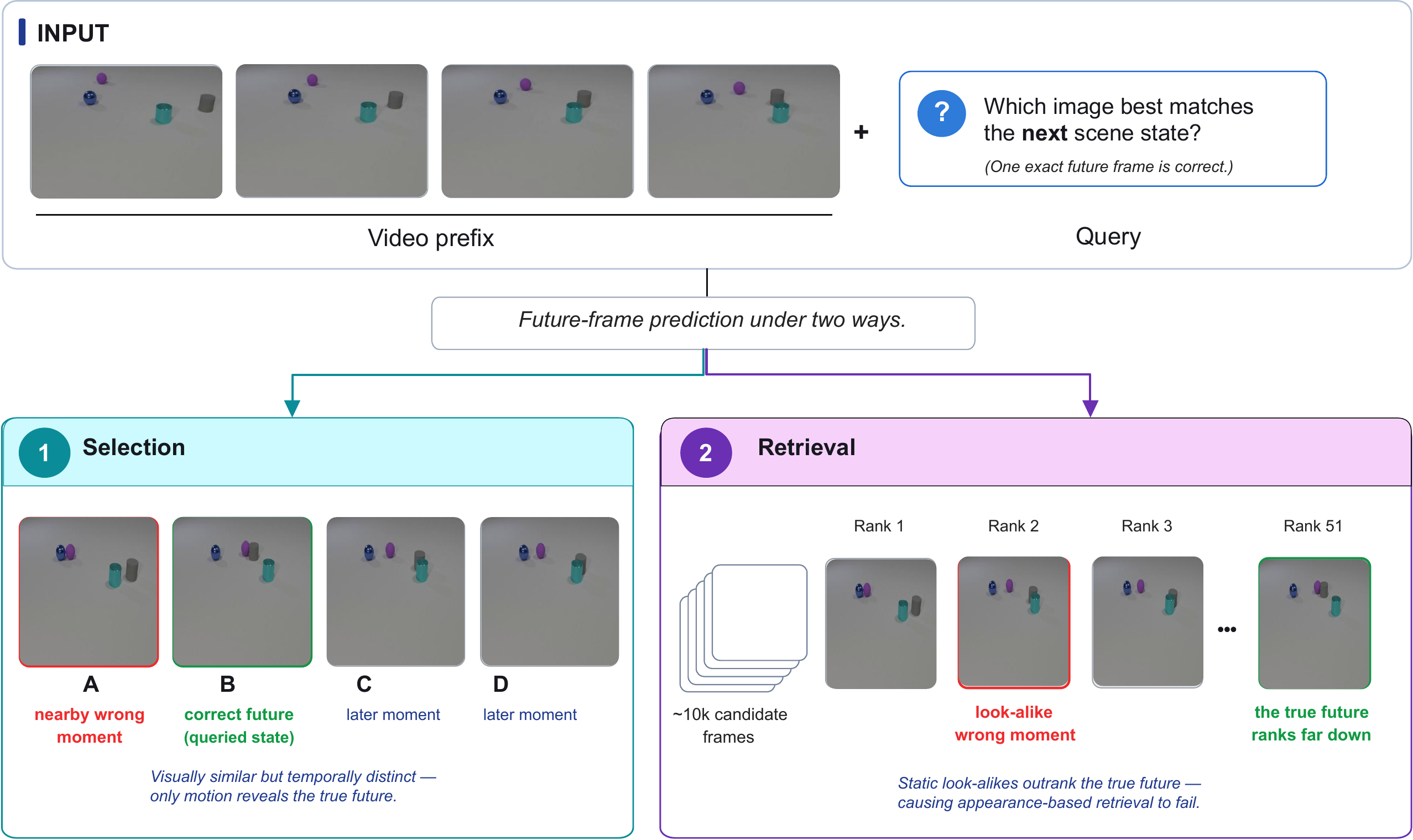}
\caption{A \bench{} example from a real benchmark instance. Selection: observed prefix plus future-oriented query, choose the correct future frame (green) among four same-scene candidates. Retrieval: rank the correct future in the \ndbimages{}-frame standard database. Scenes are simulated with the Bullet physics engine.}
\label{fig:teaser}
\end{figure*}

Existing evaluations of physical future prediction do not check this commitment.
Language-answer evaluations score a textual continuation, a one-word answer, or a
free-text forecast
\citep{lei2020more,yi2019clevrer,patel2022cripp,chang2026scattered}, but answers
such as \textit{``they will collide''} underdetermine where each object comes to
rest and may reflect what \emph{typically} happens rather than what happens in
\emph{this} scene. Generated-frame evaluations instead reward visual fidelity
rather than whether the depicted state is the correct future
\citep{li2026beyond}. What is missing is a verifiable test that scores a
committed, specific future against ground truth.

We introduce \textbf{\bench{}} (\textbf{Dyna}mic \textbf{Pix}els), a benchmark
for verifiable future-state prediction. Given a video prefix that stops before a
key event and a future-oriented query, a model must identify the specific
resulting scene, either in \textbf{selection}, where the target is chosen from
plausible candidates, or in \textbf{retrieval}, where it is found in a large
image database. Since both settings need a known-correct target and controlled
distractors, we build \bench{} on simulated scenes rather than recorded video.
Simulation gives the exact resulting state and lets us construct hard negatives
that preserve appearance while changing the dynamics, the same scene at the wrong
moment and look-alike scenes from other videos
(Figure~\ref{fig:teaser}), making surface matching and language priors less
reliable. We span multiple simulated domains
\citep{greff2022kubric,bear2021physion} to reduce dependence on any one
simulator's artifacts.

Across \bench{}, the open models and multimodal retrievers we evaluate do not
predict the specific future reliably. When the query asks for the exact state at
a specified later moment, they perform near chance: Qwen3-VL
\citep{bai2025qwen3} reaches only about 27\%, against the 25\% that random choice
among the candidates would give. They do better when a collision marks the target
moment, about 65\%, still far from reliable. This time-anchored weakness appears in
every model we test, across generative and embedding families, though matched-anchor
controls are needed to separate temporal anchoring from construction factors. In a
human validity probe, annotators identify the exact later state on 83.3\% of the
elapsed-time items of a balanced subsample while models stay near chance, so
those items leave real headroom despite the study's modest size. Retrieval tests the same commitment in an open-set setting, and the
strongest retriever we test, Qwen3-VL-Emb-8B \citep{li2026qwen3}, ranks the
correct future first only 13.3\% of the time, so the correct frame is in the
database yet rarely surfaced.

We also study physics-grounded chain-of-thought distillation, which supervises a
student with simulator-derived accounts of how each scene unfolds rather than a
teacher's unsupported rationales. It substantially improves event- and
window-anchored selection and lifts the one-second time case, and contrastive
finetuning of the retriever raises its Recall@1 from 13 to 49\%. The hardest
cases remain: selecting the state at a longer elapsed horizon and ranking the
exact future first, which leave \bench{} with remaining headroom.

We summarize our contributions as follows:
\begin{itemize}
\item We introduce \bench{}, a benchmark that makes predictive understanding
verifiable by asking a model to identify the specific future state of a physical
scene, by selection among controlled candidates or retrieval from a large database.
\item We propose physics-grounded chain-of-thought distillation, which supervises
a student with reasoning grounded in a simulator's true dynamics rather than a
teacher's generated rationales, substantially improving selection over answer-only
training.
\item We analyze current open generative models and multimodal retrievers and find
them far from solving \bench{}. They are near chance on the state at a specified
later time, a failure distillation lifts only at the shortest horizon, and they are
weak at zero-shot open-set retrieval, where targeted finetuning closes much but not
all of the gap. A human study supports the validity of the time-anchored gap.
\end{itemize}

\section{Related Work}
\label{sec:related}

\paragraph{Future prediction in language space.}
Many predictive evaluations ask for an answer in words. VLEP scores which of two
textual continuations is more likely \citep{lei2020more}, synthetic physics suites
pose predictive questions whose answers are words or labels
\citep{yi2019clevrer,patel2022cripp}, and open-ended event forecasting asks a
model to write what happens next
\citep{chang2026scattered,yu2025forestcast,li2025cllmate}. Such targets underdetermine the resulting state: a fluent continuation can name the event yet omit where each object ends up, rewarding typical outcomes over this scene's outcome. \bench{} keeps
the query in language but moves the answer into pixels, so a response is scored
against the frame the simulator actually produced.

\paragraph{World models and predictive probes.}
Recent probes report that vision-language models describe observed scenes far
better than they predict continuations
\citep{qiu2026can,li2026word,qian2026current}, and related diagnostics isolate
causal and object-state reasoning as distinct weaknesses
\citep{komanduri2025causalvlbench,nguyen2024oscar}. Simulation is a standard reference for physical prediction, in cognitive
accounts \citep{battaglia2013simulation,smith2019modeling} and rigid-body benchmarks
\citep{greff2022kubric,bear2021physion}. Generated-frame evaluations score the
fidelity of a synthesized image \citep{li2026beyond}. We score identity against the
true frame, separating families where a salient event marks the target moment
from the family where only elapsed time does.

\paragraph{Multimodal retrieval.}
Our retrieval track uses methods from composed and interleaved
multimodal retrieval, where a query combines modalities and evaluation must
resolve fine-grained distinctions \citep{song2026rethinking,tang2025modeling}, and
from multimodal embedders trained for that setting
\citep{zhang2025towards,feng2026generative}. Late-interaction and hybrid
retrievers preserve token-level correspondences that single-vector encoders
discard \citep{santhanam2022colbertv2,kim2026hybrid}, which fits frames that differ only in object position. We leave them to future work, so our claims do not depend on them.

\paragraph{Chain-of-thought distillation.}
Distilling rationales from a teacher improves small students across reasoning
tasks
\citep{hsieh2023distilling,wang2023scott,zhang2025improve,goncharov2026complexity}.
These recipes assume something that fails for physical prediction: the rationale
is sampled from a black-box teacher, so a fluent but wrong explanation supervises
the student as strongly as a correct one. We keep the format but change the source:
our rationales are derived from the simulator's record, converted into
qualitative predicates, and checked for numeric leakage, so the teacher narrates a
true account rather than inventing one.

\section{The \bench{} Benchmark}\label{sec:benchmark}

\subsection{Task formulation}

\bench{} evaluates whether a model can identify a \emph{specific future state} of a physical scene given an observed video prefix and a future-oriented query. Each instance contains a prefix video \(V_{\leq t}\), a query \(q\), and an image answer \(I^\star\) depicting the queried future state. The image answer is verified directly against simulator frames. Two protocols test this at two scales. In the \textbf{selection track} the model chooses the correct image from four candidates (chance \(25\%\)), and in the \textbf{retrieval track} it ranks the correct frame from a large database. Selection is a controlled closed-set diagnostic. Retrieval is an open-set test among many visually similar future frames.

\subsection{Construction from simulator oracles}

The core \bench{} scenes use the Bullet physics engine \citep{coumans2021pybullet} and follow the collision-scene design of CLEVRER \citep{yi2019clevrer}. We generate 10,000 collision scenes with oracle annotations of object trajectories and collision events, controlling object placement, event timing, and the exact future trajectory of every scene. Ground truth is a real simulation frame, not a human or MLLM-generated label, making each instance \textbf{verifiable} by frame identity and timestamp.

The prefix boundary is set from oracle event timestamps: for event-based items, the prefix ends a default 10 frames before the target event, preventing event leakage and forcing prediction from prior dynamics. Selection items resist appearance shortcuts with \emph{same-scene wrong-time} frames as the main distractor. These share object identities, colors, shapes, viewpoint, and background with the correct frame but come from another moment. Because all candidates are real simulator frames, no option is singled out by rendering artifacts or physical impossibility. The templates specify the temporal anchor and horizon but never name the answer image or encode the target as a text label.

\subsection{Selection track}

\begin{figure*}[t]
\centering
\includegraphics[width=\textwidth]{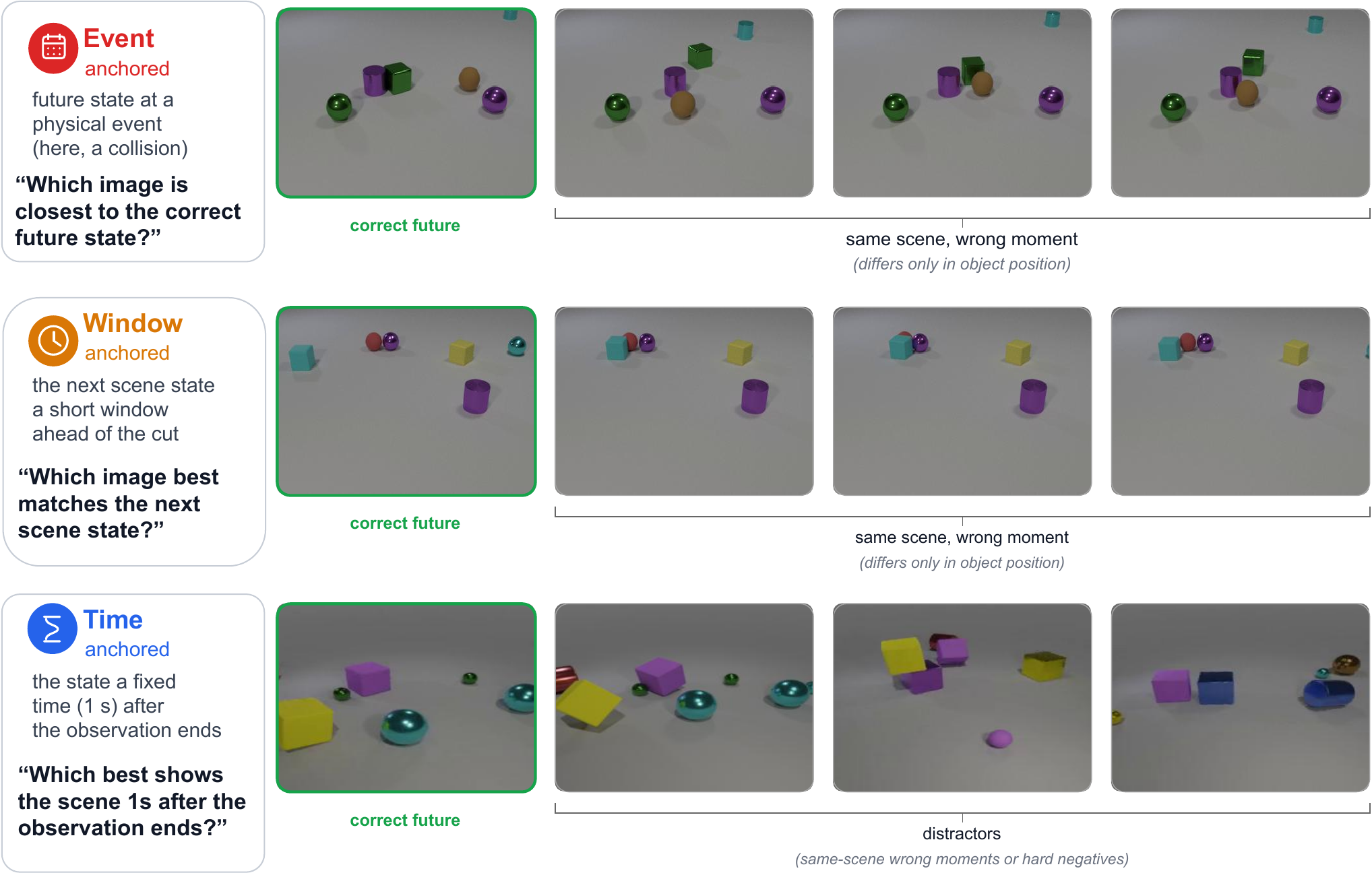}
\caption{The three selection anchors, shown on real benchmark instances. Given an observed prefix cut before the target, choose one exact future frame $t^{\star}$ (green) among four candidates. \textbf{Event-anchored}: frame right after the next collision. \textbf{Window-anchored}: state a short horizon ahead. \textbf{Time-anchored}: state a fixed time after the cut. Distractors are wrong moments from the same scene or hard negatives from a different scene.}
\label{fig:anchors}
\end{figure*}

The selection track has three question families, defined by how time anchors the target future state. Appendix~\ref{app:families} gives worked examples and construction variants, and Appendix~\ref{app:subtypes} gives subtype counts.

\textbf{Event-anchored.} The prefix stops just before a collision, and the model selects the frame that shows the scene \emph{right after that collision}. The anchor is event-relative, so the task requires predicting the outcome of the interaction rather than choosing any plausible later frame.

\textbf{Time-anchored.} The model observes a short prefix and must identify the state at \emph{a precise later time}, 1 or 2 seconds ahead. No event need occur at that instant, so the family probes temporal precision rather than event recognition. Candidates a few frames apart can conflate forward prediction with sub-second timestamp discrimination, and Section~\ref{ssec:human} checks this.

\textbf{Window-anchored.} The model observes a brief activity window and must identify the state \emph{a short horizon into the future}. Every window item uses one shared query, so items differ in the physical situation rather than the wording. The future window may contain no event, one collision, an object entering or leaving view, or several events.

Distractor sources follow the temporal anchor. Time-anchored and window-anchored items use same-scene wrong-time distractors, while event-anchored items also include a cross-scene distractor variant that tests whether the model tracks the correct scene rather than only the correct moment. We release a fixed split of 9,368 training and 2,208 test questions over 387 distinct test scenes. Each item has four candidates and exactly one correct image. Table~\ref{tab:dynapix_stats} summarizes the split, retrieval pools, and transfer suites. Window-anchored items make up 72\% of the test split, so we report per-family accuracy throughout and treat any aggregate over the three families as a reference number only.

\begin{table}[t]
\centering
\small
\setlength{\tabcolsep}{4pt}
\begin{tabular}{@{}lrr@{}}
\toprule
Subset & Count & Frames \\
\midrule
\multicolumn{3}{@{}l}{\emph{Selection} (train\,/\,test, four candidates)} \\
\quad Event-anchored & 1{,}151\,/\,249 & -- \\
\quad Time-anchored & 1{,}619\,/\,374 & -- \\
\quad Window-anchored & 6{,}598\,/\,1{,}585 & -- \\
\quad Total & 9{,}368\,/\,2{,}208 & -- \\
\midrule
\multicolumn{3}{@{}l}{\emph{Retrieval} (queries\,/\,database frames)} \\
\quad Future-state pool & 22{,}500 & 30{,}000 \\
\quad Counterfactual pool & 18{,}225 & 26{,}995 \\
\quad Standard test & 2{,}002 & 10{,}000 \\
\midrule
\multicolumn{3}{@{}l}{\emph{Transfer, zero-shot} (test items)} \\
\quad MoVi-A (event\,/\,time) & 327\,/\,1{,}198 & -- \\
\quad MoVi-A total & 1{,}525 & -- \\
\quad Physion (event\,/\,time) & 583\,/\,1{,}612 & -- \\
\quad Physion total & 2{,}195 & -- \\
\bottomrule
\end{tabular}
\caption{\bench{} statistics. Selection uses four candidates per item, a frozen split, and 387 distinct test scenes. Retrieval lists the standard test set and its source pools.}
\label{tab:dynapix_stats}
\end{table}

\subsection{Retrieval track}

\begin{figure*}[t]
\centering
\includegraphics[width=\textwidth]{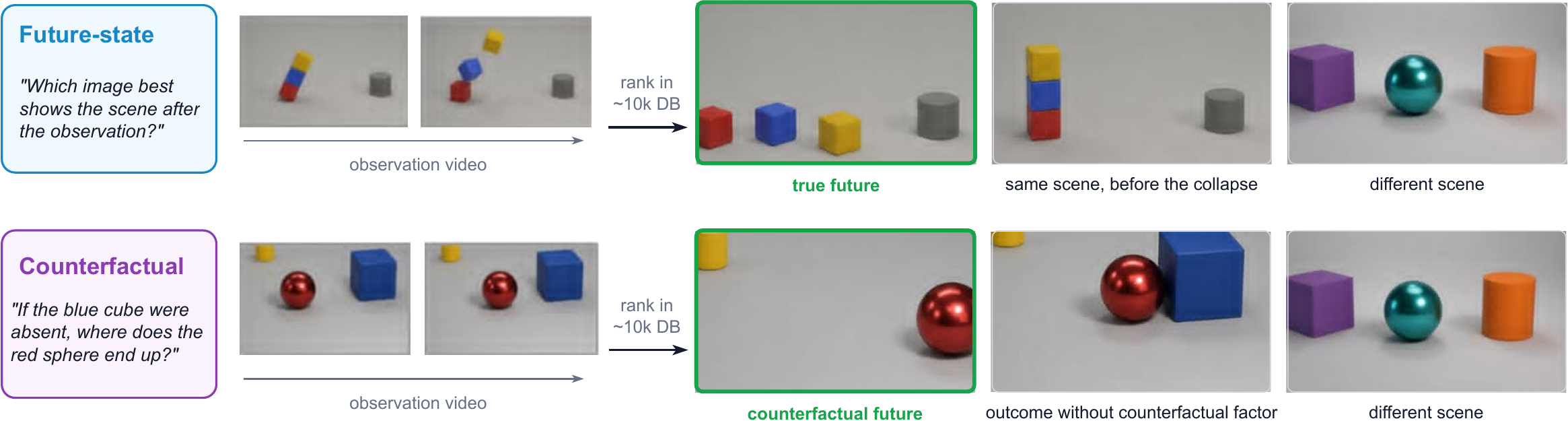}
\caption{The retrieval track of \bench{} (illustrative). From the observation video, a method ranks the true future frame (green) among a database of about \ndbimages{} frames. \textbf{Future-state} queries ask for the scene after the observation: the true future is the outcome the dynamics force, here the toppling stack settled flat, while distractors show the scene before the collapse or a different scene. \textbf{Counterfactual} queries remove a named object and ask for the resulting motion: without the blue cube the red sphere rolls on, whereas the factual outcome keeps the cube and the sphere stops against it.}
\label{fig:retrieval}
\end{figure*}

The retrieval track replaces closed-set choice with open-set ranking. Given the same prefix and query, the model ranks the correct future frame in a database, so the answer must be found among thousands of look-alikes rather than a small candidate set (Figure~\ref{fig:retrieval}). The two query families (Table~\ref{tab:dynapix_stats}) are \textbf{future-state}, which varies the observation length and prediction horizon, and \textbf{counterfactual}, which adds an intervention by asking where an object would be if another were absent. The standard evaluation uses \nretqueries{} queries over \ndbimages{} frames, and Appendix~\ref{app:retrieval} gives further detail.

\subsection{Transfer suites}

To measure out-of-simulator transfer, we build zero-shot suites with the same oracle recipe on MoVi-A, a Kubric rigid-body setting, and Physion \citep{greff2022kubric,bear2021physion}. Both suites contribute event- and time-anchored items, with the Physion items drawn from three regimes: A1 contact, A2 motion and A3 landing. Counts are in Table~\ref{tab:dynapix_stats} and construction details are in Appendix~\ref{app:transfer}.

\section{How Far Are VLMs from Predictive Understanding?}
\label{sec:eval}

We use \bench{} to test whether current models can identify a
specific future state and whether performance depends on target anchoring: a
salient physical event, elapsed time alone, or a short future window. We report
the selection track in Section~\ref{ssec:eval-selection} and the open-set
retrieval track in Section~\ref{ssec:eval-retrieval}. All model numbers are
zero-shot before \bench{} training. Section~\ref{sec:method} tests whether
physics-grounded supervision closes the gap.

\subsection{Models and metrics}

A selection item admits two solver types. A \emph{generative} model reads the
prefix, query, and four candidates and writes a choice. A \emph{similarity}
scorer embeds the query and candidates and picks the nearest candidate without
generating a future state. The comparison separates prediction from appearance
matching. \rev{As generative models, we use Qwen3-VL-2B and
Qwen3-VL-8B in instruct mode, Qwen3-VL-8B in thinking mode
\citep{bai2025qwen3}, and the larger open models Qwen3.5-27B and Qwen3.6-27B,
which test whether scale alone supplies the missing ability. As similarity
scorers, we use CLIP \citep{radford2021learning}, SigLIP~2
\citep{tschannen2025siglip}, LanguageBind \citep{zhu2024languagebind}, VLM2Vec
\citep{jiang2025vlm2vec}, and Qwen3-VL-Embedding at 2B and 8B
\citep{li2026qwen3}, each a dual encoder applied to both the selection and
retrieval tracks.}
For thinking models, unparseable outputs count as wrong, and we report
parseability because a model that cannot commit has not answered
(Appendix~\ref{app:protocol}).

For retrieval, each method ranks every database frame against the query. Direct
embedders compare the query and each frame in one step.
\rev{Two-stage pipelines instead generate a textual future prediction and match
it to the database through a frozen text-image encoder. Our strongest such
pipeline generates a chain of thought with Qwen3-VL-8B-Instruct, while two
weaker \emph{predict} variants generate a short outcome description with
Qwen3-VL-2B and match it to generated frame captions or the frames themselves.
Since the encoder is identical across two-stage rows, only the generative stage
differs.} We report Recall@$k$, mean reciprocal rank (MRR), and the median rank
of the true future among 10,000 frames.

\subsection{Selection: event-anchored items are solved, time-anchored ones are not}
\label{ssec:eval-selection}

\begin{table}[t]
\centering
\small
\setlength{\tabcolsep}{4pt}
\resizebox{0.9\columnwidth}{!}{%
\begin{tabular}{@{}lccc@{\hskip 10pt}c@{}}
\toprule
 & Event & Time & Window & Mean \\
\midrule
Chance & 25.0 & 25.0 & 25.0 & 25.0 \\
Human$^\ddagger$ & 58.3 & 83.3 & 73.9 & 71.8 \\
\midrule
\multicolumn{5}{@{}l}{\emph{Similarity scorers}} \\
\quad CLIP & 44.6 & 27.0 & 40.7 & 37.4 \\
\quad LanguageBind & 18.9 & 21.7 & 24.1 & 21.5 \\
\quad SigLIP 2 & 65.9 & 20.6 & 52.7 & 46.4 \\
\quad VLM2Vec & \textbf{71.1} & 24.1 & \textbf{55.6} & \textbf{50.2} \\
\quad Qwen3-VL-Emb-2B & 63.1 & 21.4 & 44.7 & 43.1 \\
\quad Qwen3-VL-Emb-8B & 57.0 & 22.7 & 52.7 & 44.1 \\
\midrule
\multicolumn{5}{@{}l}{\emph{Open generative VLMs}} \\
\quad Qwen3-VL-2B (instruct) & 28.5 & 24.6 & 35.3 & 29.5 \\
\quad Qwen3-VL-8B (instruct) & 64.7 & \textbf{27.3} & 42.0 & 44.7 \\
\quad Qwen3-VL-8B (thinking)$^\dagger$ & 65.1 & 18.7 & 39.3 & 41.0 \\
\quad Qwen3.5-27B & 25.8 & 21.6 & 26.6 & 24.7 \\
\quad Qwen3.6-27B & 26.1 & 20.3 & 27.1 & 24.5 \\
\bottomrule
\end{tabular}}
\caption{Zero-shot selection accuracy (\%) on the \bench{} test set. Columns are the three \textbf{per-family} scores Event, Time, Window, and their unweighted Mean. Test mix: 72\% window-anchored, so the per-family scores are primary. Chance is 25\%. Bold marks the best model result per family. $^\dagger$Unparsed outputs are counted as wrong throughout. Parse rates in reasoning mode are 66.4\% for the thinking base, 89.9\% for Qwen3.5-27B and 81.1\% for Qwen3.6-27B, so the 27B rows also mix errors with non-answers. $^\ddagger$Human is the three-annotator majority vote on a balanced cross-simulator subsample (Section~\ref{ssec:human}), read as a validity probe, so its per-family scores are comparable but its item mix differs from the test set.}
\label{tab:selection}
\vspace{-10pt}
\end{table}

Table~\ref{tab:selection} reports per-family accuracy. We treat these as primary
and the mean as reference, because the test set is 72\% window-anchored.
A pooled score would be dominated by window and hide failure on the time family.
Accuracy separates sharply by family. With a salient collision marking
the target moment, \ev{the best generative model reaches 65.1\%, and VLM2Vec
reaches 71.1\%, both well above the 25\% chance rate}. With elapsed time alone,
\ev{every model is near chance, with a best of 27.3\%}. The contrast holds within
Qwen3-VL-8B instruct: 64.7\% on event and 27.3\% on time, so the pattern is not a
quirk of one architecture. Whether it reflects temporal anchoring itself or other
differences between the families is taken up in Section~\ref{ssec:human}.

Neither scale nor explicit reasoning removes the gap. Thinking mode raises event
from 64.7\% to 65.1\% but lowers time from 27.3\% to 18.7\%. The larger open
models are the clearest scale test: Qwen3.5-27B and Qwen3.6-27B sit within a few
points of chance on every family, so scale alone does not confer the missing
ability. The similarity scorers extend the pattern. VLM2Vec never generates a
future state, yet posts the highest mean at 50.2\%,
driven by event and window items where appearance is partly informative, while it
too falls to 24.1\% on time. The unweighted mean can thus be led by appearance
matching. The per-family view exposes the time-anchored failure \bench{} is
designed to surface: every evaluated similarity scorer is within a few points of
chance on time, so appearance matching does not recover a moment specified only
by elapsed time.

\subsection{Human performance and temporal validity}
\label{ssec:human}

Near-chance accuracy on time-anchored items has two readings: models lack the
capability, or the items are ill posed for anyone. We use a human study as a
validity probe. Three annotators judged a balanced cross-simulator subsample
with randomized candidate order and balanced answer position, so chance is
25\%. Of 59 unique items, 10 were seen by one annotator, 30 by two, and 19 by
all three, for 127 judgments. Because depth varies, the majority vote aggregates
the available judgments rather than a uniform three-way vote.

The time family supports validity. Human majority accuracy is 83.3\% versus 27.3\%
for the best model, with annotators at 83.3\%, 100.0\%, and 83.3\%. Thus
near-chance model performance on elapsed-time queries is not due to unanswerable
time items or indistinguishable adjacent future states, and shows headroom for
selecting a later state specified only by elapsed time.

On event-anchored items humans are not above models. The majority vote is 58.3\%
versus 71.1\% for VLM2Vec and 65.1\% for the best Qwen3-VL-8B setting, and
annotators at 58.3\%, 40.0\%, and 63.6\%. Event is slowest for humans, median
39.9s versus 15.8s for time and 20.3s for window. Thus humans find time easiest
and event hardest, the reverse of models, which are near chance on time and much
stronger on event.

This reversal argues against event items being simply easier for any observer
due to obvious static post-collision targets: then humans should find event
easy, but they do not. Still, this is not a controlled test of the
construction-confound concern: the families differ beyond the anchor in possible
horizon, distractor, and target-state properties, and it does not rule out
model-specific static cues, candidate-only artifacts, or post-collision
appearance strategies that ignore the prefix.

Agreement is fair. Fleiss' $\kappa$ is 0.366 on the 19 items seen by all three
annotators. We treat it as a validity probe, not a human ceiling or causal
isolation of temporal anchoring. The strongest conclusion is that elapsed-time
items are human-answerable while all evaluated models remain near chance. A
matched-anchor experiment with identical prefix, target, horizon, and
candidates, differing only in event versus elapsed-time anchors, plus
candidate-only and shuffled-prefix controls, would settle it.

\subsection{Retrieval: the true future is rarely ranked first}
\label{ssec:eval-retrieval}

\begin{table}[t]
\centering
\small
\setlength{\tabcolsep}{3.5pt}
\resizebox{\columnwidth}{!}{%
\begin{tabular}{@{}lccccc@{}}
\toprule
Method & R@1 & R@10 & R@50 & MRR & Med.\ rank \\
\midrule
\multicolumn{6}{@{}l}{\emph{Direct embedders}} \\
\quad CLIP & 1.9 & 8.0 & 20.4 & 0.041 & 404 \\
\quad LanguageBind & 1.0 & 6.9 & 19.4 & 0.031 & 388 \\
\quad SigLIP 2 & 7.3 & 39.4 & 60.6 & 0.171 & 23 \\
\quad VLM2Vec-V2 & 12.7 & 62.0 & 81.1 & 0.272 & \textbf{6} \\
\quad Qwen3-VL-Emb-2B & 10.6 & 49.1 & 71.5 & 0.224 & 11 \\
\quad Qwen3-VL-Emb-8B & \textbf{13.3} & \textbf{64.2} & \textbf{81.3} & \textbf{0.283} & \textbf{6} \\
\midrule
\multicolumn{6}{@{}l}{\emph{Two-stage predict-then-match} (matching encoder frozen)} \\
\quad Qwen3-VL-8B CoT $\rightarrow$ text & 2.1 & 11.4 & 26.6 & 0.055 & 252 \\
\quad Qwen3-VL-2B predict $\rightarrow$ caption & 1.0 & 6.7 & 18.0 & 0.031 & 477 \\
\quad Qwen3-VL-2B predict $\rightarrow$ image & 0.2 & 2.4 & 7.6 & 0.012 & 985 \\
\bottomrule
\end{tabular}}
\caption{Zero-shot retrieval, standard test set: 2{,}002 queries, 10{,}000-frame database. Recall@$k$ and MRR are higher-is-better. Median rank is lower-is-better. Chance R@1 is 0.01\%. Bold marks the best result per column.}
\label{tab:retrieval}
\vspace{-10pt}
\end{table}

Retrieval is stricter: the correct future sits in a database of about 10k
look-alikes and must outrank them. Even the strongest embedder, Qwen3-VL-Emb-8B,
places the true future first only 13.3\% of the time, at an MRR of 0.283 and a
median rank of 6. The correct frame is usually near the top rather than first, so
a system that acts on one retrieved image is right 13.3\% of the time. The
two-stage predict-then-match pipelines are weaker, because generating a future
description and matching it to pixels can lose the fine spatial detail that
separates one moment from the next. Across zero-shot methods here, open-set
localization of the exact future remains unsolved, and Section~\ref{sec:method}
asks how far supervised adaptation closes it.

\section{Physics-Grounded CoT Distillation}
\label{sec:method}

Section~\ref{sec:eval} leaves one supervision question: can training rationales
tied to the true future state improve a model? Standard chain-of-thought
distillation trains the student on a teacher's own rationales
\citep{hsieh2023distilling,wang2023scott}. For physical prediction, those
rationales can be scene-mismatched guesses. We ground each rationale in the
simulator record, so the supervision is physically correct by construction.

\subsection{Method}
\label{ssec:method-pipeline}

The pipeline grounds each training scene in three steps. \textbf{Oracle to
predicates:} we read the simulator's exact trajectories and collision events and
convert them into qualitative predicates: direction and relative speed of each
object, which pairs are on a collision course, and each object's region at the
target moment. A validator strips all raw coordinates, timestamps, and numeric
values, so rationales use only test-time terms. \textbf{Teacher narration:} a
Qwen3-32B teacher phrases the predicates into a chain of thought that parses the
scene, analyzes motion, predicts the configuration, and compares it against the
candidates (Appendix~\ref{app:rationales}). \textbf{Student training:} we
fine-tune Qwen3-VL at 8B and 2B with LoRA on pairs mapping the prefix and
question to the rationale and answer (Appendix~\ref{app:impl}). At inference the
student sees only pixels and the question, so the oracle is unavailable.

\textbf{Retrieval adaptation.} We adapt both retrieval families. In the two-stage
pipeline, we replace the generative stage, Qwen3-VL-8B-Instruct, with the
distilled 8B student and freeze the matching encoder, isolating grounded
reasoning. For the direct family, we contrastively fine-tune Qwen3-VL-Emb-8B, the
strongest zero-shot retriever, with LoRA on the retrieval pool of 22,500 queries,
using in-batch and same-scene hard negatives. Both train on the training pool only
and are evaluated on the standard test set.

\subsection{In-domain results}
\label{ssec:method-indomain}

\begin{table}[t]
\centering
\small
\setlength{\tabcolsep}{4pt}
\resizebox{\columnwidth}{!}{%
\begin{tabular}{@{}lccc@{\hskip 10pt}cc@{}}
\toprule
\multicolumn{6}{@{}l}{\emph{(a) Selection, per family (Qwen3-VL-8B)}} \\
 & Event & Time & Window & Macro & Micro \\
\midrule
Zero-shot (base) & 64.7 & 27.3 & 42.0 & 44.7 & 42.1 \\
Answer-only SFT & 73.9 & 36.6 & 61.1 & 57.2 & 58.4 \\
Physics-grounded CoT & \textbf{92.4} & \textbf{45.4} & \textbf{96.7} & \textbf{78.2} & \textbf{87.5} \\
\midrule
\multicolumn{6}{@{}l}{\emph{(b) Retrieval}} \\
 & R@1 & R@10 & R@50 & MRR & Med. \\
\midrule
Two-stage (Qwen3-VL-8B $\rightarrow$ text) & 2.1 & 11.4 & 26.6 & 0.055 & 252 \\
\quad + CoT-distilled student & 16.0 & 61.3 & 78.0 & 0.305 & 6 \\
Direct (Qwen3-VL-Emb-8B) & 13.3 & 64.2 & 81.3 & 0.283 & 6 \\
\quad + contrastive finetune & \textbf{48.8} & \textbf{92.0} & \textbf{96.8} & \textbf{0.652} & \textbf{2} \\
\bottomrule
\end{tabular}}
\caption{Physics-grounded CoT distillation. \textbf{(a)} CLEVRER selection accuracy per family. \textbf{(b)} Retrieval on 2{,}002 queries over 10{,}000 frames. The distilled student replaces the two-stage pipeline's generative stage, and the contrastive row finetunes the direct embedder. Bold marks the best result in each column of each panel.}
\label{tab:distill}
\vspace{-10pt}
\end{table}

Table~\ref{tab:distill} shows uneven CLEVRER selection gains for Qwen3-VL-8B.
Distillation nearly solves window, from 42.0 to 96.7, and strongly lifts event,
from 64.7 to 92.4, because both families reward collision-outcome reasoning
supplied by the grounded rationale. Answer-only training reaches only 61.1 on
window and 73.9 on event, so most of the gain comes from rationale content rather
than answer supervision alone (Appendix~\ref{app:ablation}). 

Time-anchored accuracy is the weakest family under every regime. Distillation
lifts it from 27.3 to 45.4, above answer-only fine-tuning at 36.6, but the gain
comes from the near horizon: the 1-second case rises to 64.0 while the 2-second
case reaches only 37.6, from base rates of 28.8 and 26.6. A qualitative rationale
supports the one-second case, but performance decays as the horizon lengthens and
the target drifts from observed landmarks. The 2-second case remains close to the
failure pattern in Section~\ref{sec:eval}. Distillation helps most when a landmark
or short reach anchors the moment. Longer-horizon anchoring remains open.

\subsection{Transfer to unseen simulators}
\label{ssec:method-transfer}

Transfer to MoVi-A and Physion is partial (Section~\ref{sec:benchmark}). Applied
zero-shot, the CLEVRER-trained student improves overall selection accuracy
(answer mode) from 39.3 to 48.1 on MoVi-A and from 17.5 to 48.4 on Physion. Both
gains are below the in-domain rise from 42.1 to 86.7 measured in the same answer
mode. The same family split persists: on MoVi-A the improvement is concentrated
in event-anchored items, from 72.8 to 89.3, while time-anchored items barely move,
from 30.2 to 36.8. Transfer again fails on the elapsed-time case.

\subsection{Retrieval with distilled reasoning}
\label{ssec:method-retrieval}

With the matching encoder frozen, swapping the distilled 8B student into the
generative stage of the two-stage pipeline raises MRR from 0.055 to 0.305 and cuts
the median rank of the true future from 252 to 6 (Table~\ref{tab:distill}).
Because only the generative stage changes, the gain reflects the grounded
rationale rather than a stronger matcher, and exceeds the zero-shot direct
embedder it trailed, 0.305 against 0.283 MRR.

Two-stage adaptation leaves the stronger direct-retrieval family untrained, so we
also contrastively fine-tune Qwen3-VL-Emb-8B
(Section~\ref{ssec:method-pipeline}), reported in Table~\ref{tab:distill}(b). This
finetune is our strongest retriever: Recall@1 rises from 13.3 to 48.8, MRR from
0.283 to 0.652, Recall@50 from 81.3 to 96.8, and the median rank from 6 to 2.
These gains show representation headroom, though even the finetuned embedder ranks
the exact future first fewer than half the time.

\section{Conclusion}
\label{sec:conclusion}

We introduce \bench{}, a benchmark that tests whether vision-language models
identify the exact future frame of a physical scene. With simulator oracles,
\bench{} has verifiable targets and appearance-matched distractors. We also
propose physics-grounded chain-of-thought distillation, where the teacher narrates
the simulator record rather than inventing it. This supervision nearly solves the
event- and short-horizon families, and a contrastively finetuned embedder closes
much of, but not all, the open-set retrieval gap. Temporal precision remains the
main failure mode: models stay near chance when elapsed time alone fixes the target
moment, and distillation helps only at the shortest horizon. The human study places
annotators far above models on elapsed-time items, leaving headroom for future
work on prediction with temporal precision. The event-anchored comparison needs
controlled follow-up, and a matched-anchor design with candidate-only controls is
the natural next experiment, holding every factor but the anchor fixed.

\clearpage
\section*{Limitations}

\bench{} gives simulator-verifiable, frame-exact targets for future-state
prediction. This verifiability restricts scope.

\begin{itemize}
\item \textbf{The families differ in more than their anchor.} This is the
limitation that most constrains our interpretation. Event-anchored and
time-anchored items differ not only in what fixes the target moment but also in
prediction horizon, distractor sourcing, and how visually distinctive the target
state is, since a post-collision configuration is salient in a way that an
arbitrary point on a trajectory is not. We therefore report a family-level
difference and cannot attribute it to temporal anchoring alone. Two observations
argue against the simplest competing account, that event items are easier for any
observer: annotators are at or below models on event items while far above them on
time items, and they take longest on event. Neither is a controlled manipulation.
Settling this needs matched pairs that share a prefix, target frame, horizon and
candidate set while varying only whether the query names an event or an elapsed
time, together with candidate-only and shuffled-prefix controls that test how far
the target can be identified without the observed dynamics. We did not run those
experiments and do not claim the mechanism.

\item \textbf{Synthetic scope.} The data are synthetic. This makes
frame-identity ground truth auditable, but leaves real-world transfer untested.
Future work should rebuild the oracle from tracks, robot logs, or instrumented
video and test whether the temporal-anchoring gap survives clutter, camera
motion, and imperfect perception.

\item \textbf{Temporal resolution.} Frame-exact selection could mix prediction
with sub-second timestamp discrimination. The human study tests this directly:
annotators reach 83\% on time-anchored items, so nearby futures are
distinguishable at this resolution and model failure is not an artifact of
impossible timing. The balanced subsample is modest, so it validates the family
but does not fix the granularity at which the distinction becomes hard. A larger
calibrated study should map that boundary.

\item \textbf{Time-anchored failure.} The time-anchored regime remains the
hardest family after scaling, explicit reasoning, and oracle-grounded
distillation. The 1s and 2s split shows that failure grows with elapsed time:
distillation lifts the 1s case to 64.0 but the 2s case to only 37.6. We
characterize this horizon dependence but do not identify its cause. Future work
should separate error accumulation in predicted dynamics from failure to convert
elapsed time into displacement.

\item \textbf{Design coverage.} The oracle uses a designer-chosen predicate
vocabulary. If that vocabulary drives the distillation gain, measurements may
reflect phrasing rather than true-record grounding. Our conclusions therefore
concern predicate-grounded diagnostics, not unrestricted physical inference. The
retrieval database is single-domain, and finetuning is limited to Qwen-family
students. Future work should vary these factors separately.

\item \textbf{Retrieval coverage.} We adapt both retrieval families to the
domain, the generative two-stage pipeline and the direct embedder, but only within
single-vector architectures. Late-interaction and multi-vector retrievers preserve
token-level correspondences discarded by a single-vector encoder
\citep{santhanam2022colbertv2,kim2026hybrid}. They remain untested for cases
where the target differs from its distractors in a small frame region. We report
the model set we could run completely rather than a broader panel of partial
results.
\end{itemize}

\nocite{*}
\bibliography{custom}
\clearpage

\appendix

\section{Selection Subtypes}\label{app:subtypes}

This appendix defines the selection subtypes. Each family uses one fixed
question, stated below with a worked example. The subtypes within a family share
that question and differ only in how the item is constructed. Each table
therefore lists the construction axes and the verified train and test counts.
Every axis is explained in the text above its table. Frame ranges are inclusive
and each video has 128 frames.

\subsection{Event-anchored}

The question is \emph{``Which image is closest to the correct future state of the whole
scene?''} As a worked example, the model observes frames 10--30, where a
gray and a yellow cylinder are on a collision course, then selects the frame
just after they collide, frame 40. The distractors are the same scene at other
moments.

Three axes generate the seven subtypes of Table~\ref{tab:event_subtypes}.
\textbf{Framing} is whether the collision falls after the observation
(\emph{future}, the prefix ends 10 or 5 frames before it) or exactly at the
observation end (\emph{contact}, the target is 10 frames later).
\textbf{Distractors} is the source of the three wrong options: another scene
(cross-scene), the same scene at offset frames, or the same scene selected by an
object-match score. \textbf{Gap} is the number of frames between the prefix end
and the collision.

\begin{table}[t]
\centering\small
\setlength{\tabcolsep}{5pt}
\begin{tabular}{@{}llcrr@{}}
\toprule
Framing & Distractors & Gap & Train & Test \\
\midrule
Future  & cross-scene     & 10 & 165 & 35 \\
Future  & same, offset    & 10 & 163 & 35 \\
Future  & same, obj-match & 10 & 165 & 35 \\
Future  & same, obj-match &  5 & 162 & 38 \\
Contact & cross-scene     & 10 & 165 & 35 \\
Contact & same, offset    & 10 & 166 & 36 \\
Contact & same, obj-match & 10 & 165 & 35 \\
\midrule
\multicolumn{3}{@{}l}{Total} & 1{,}151 & 249 \\
\bottomrule
\end{tabular}
\caption{The seven event-anchored subtypes and their counts.}
\label{tab:event_subtypes}
\end{table}

\subsection{Time-anchored}

The question is \emph{``Watch the 1-second observation video. Which image best
shows what the scene will look like $k$ seconds after the observation ends?''},
with $k$ either one or two. As a worked example, the model observes a one-second clip,
frames 0--25, then selects the frame two seconds later, frame 78. The distractors
are the same scene at other times. The only axis is \textbf{Horizon}, the offset
$k$ that the question asks about. Table~\ref{tab:time_subtypes} gives both
subtypes.

\begin{table}[t]
\centering\small
\setlength{\tabcolsep}{7pt}
\begin{tabular}{@{}lrr@{}}
\toprule
Horizon & Train & Test \\
\midrule
1 second  & 555     & 111 \\
2 seconds & 1{,}064 & 263 \\
\midrule
Total & 1{,}619 & 374 \\
\bottomrule
\end{tabular}
\caption{The two time-anchored subtypes and their counts.}
\label{tab:time_subtypes}
\end{table}

\subsection{Window-anchored}

The question is \emph{``Which image best matches the next scene state?''}
As a worked example, the model observes frames 90--109, then selects frame 125,
by which point one collision has occurred. The distractors are the same scene at
other times.

Four axes generate the subtypes. \textbf{Group} is the future-event structure.
G1 has no future event. G2 has one future collision. G3 has an occlusion change
during the observation. G4 has multiple future events. \textbf{Horizon} is 8, 16,
or 24 frames ahead. \textbf{Difficulty} is easy or hard, where hard keeps the
distractors closest in appearance to the target. \textbf{Family} sets how the
target and the distractors are chosen and does not change the question. Family A
uses the whole scene. Family B uses a collision and appears only when a
collision exists. Family C uses a single object and is excluded from G4. The 47
subtypes are the valid, populated combinations of these axes.
Table~\ref{tab:window_group_counts} defines the groups and
Table~\ref{tab:window_group_family_counts} gives the counts by group and family.

\begin{table}[t]
\centering\small
\setlength{\tabcolsep}{5pt}
\begin{tabular}{@{}lp{3.4cm}rr@{}}
\toprule
Group & Future-event structure & Train & Test \\
\midrule
G1 & no future event                   & 1{,}967 & 412 \\
G2 & one future collision               & 2{,}300 & 680 \\
G3 & occlusion change in observation    & 1{,}578 & 324 \\
G4 & multiple future events             & 753     & 169 \\
\midrule
\multicolumn{2}{@{}l}{Total} & 6{,}598 & 1{,}585 \\
\bottomrule
\end{tabular}
\caption{The four window-anchored groups and their counts.}
\label{tab:window_group_counts}
\end{table}

\begin{table}[t]
\centering\small
\setlength{\tabcolsep}{5pt}
\begin{tabular}{@{}lrrrr@{}}
\toprule
 & A & B & C & Total \\
\midrule
G1 & 178 & --  & 234 & 412 \\
G2 & 269 & 266 & 145 & 680 \\
G3 & 173 & --  & 151 & 324 \\
G4 &  84 &  85 & --  & 169 \\
\midrule
Total & 704 & 351 & 530 & 1{,}585 \\
\bottomrule
\end{tabular}
\caption{Window-anchored test counts by group and family. A dash marks an
invalid family for that group. Each populated cell is split further over three
horizons (8, 16, 24 frames) and two difficulty levels, giving 47 subtypes in
all.}
\label{tab:window_group_family_counts}
\end{table}

\section{Example Gallery}\label{app:gallery}

\begin{figure*}[t]
\centering
\includegraphics[width=\textwidth]{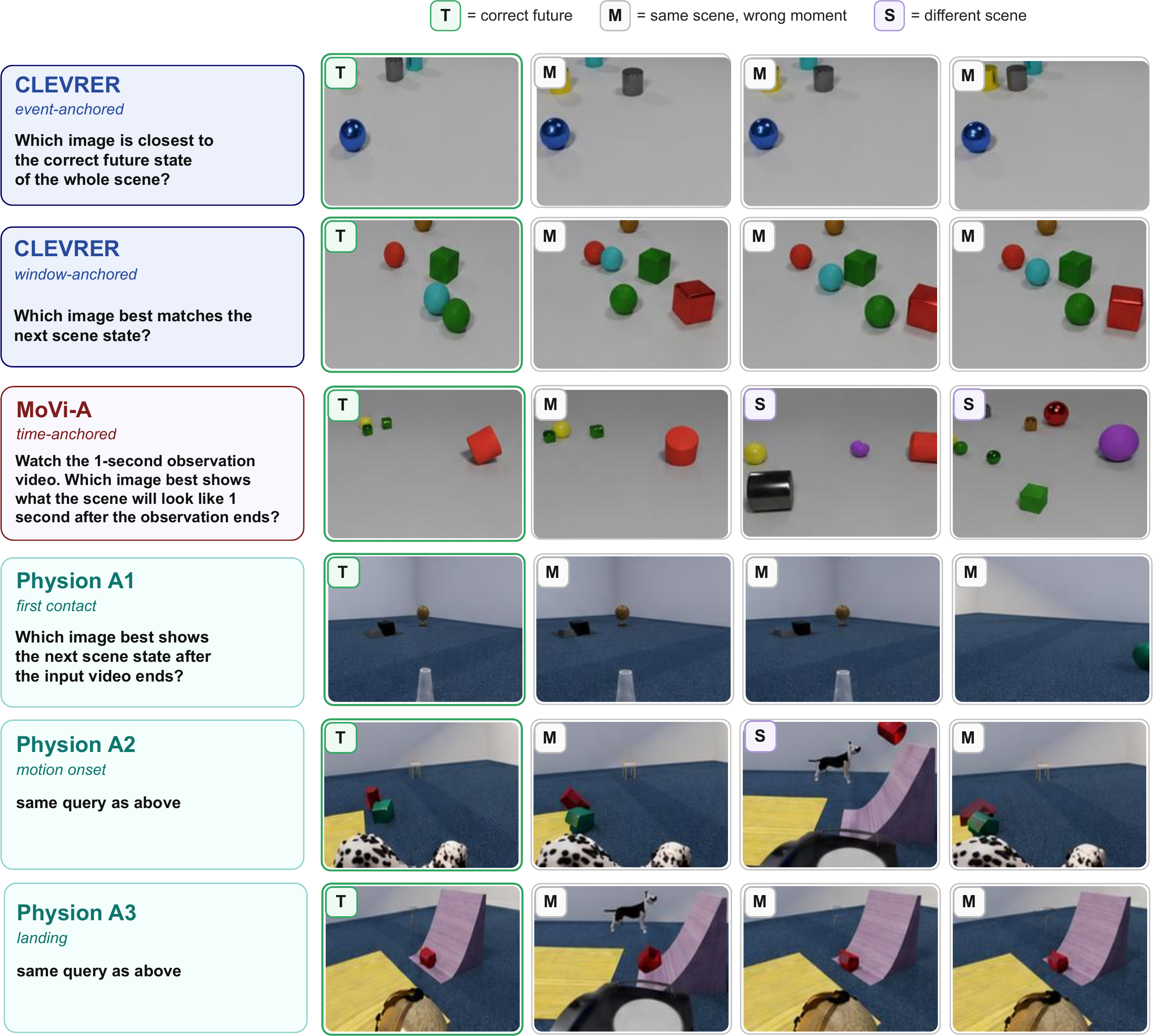}
\caption{Six \bench{} items, one per row. \textbf{CLEVRER} event-anchored and
window-anchored, \textbf{MoVi-A} time-anchored, and the three \textbf{Physion}
regimes, first contact (A1), motion onset (A2), and landing (A3). The correct
future is outlined in green. Distractors are the same scene at a wrong moment,
except in the MoVi-A row where two are drawn from different scenes.}
\label{fig:gallery}
\end{figure*}

Figure~\ref{fig:gallery} shows six items, two from each core selection family and
one from each Physion regime, chosen so that no instance repeats those in
Figures~\ref{fig:teaser} and~\ref{fig:anchors}. The gallery makes three things
visible that the main paper states but cannot show compactly. First, the same
oracle recipe produces items in three simulators with entirely different
renderers and object inventories, from CLEVRER primitives to TDW rooms containing
furniture and animals. Second, the distractor design is visible per family: the
CLEVRER and Physion rows draw every distractor from the same scene at a wrong
moment, whereas the MoVi-A row mixes one same-scene wrong moment with two
cross-scene hard negatives. Third, and most importantly, within every row the
candidates are appearance-matched, so nothing in the pixels except object
position separates the correct future from the distractors.

\section{Worked Examples of the Selection Families}\label{app:families}

Section~\ref{sec:benchmark} defines the three selection families compactly. We
give the worked examples and construction variants here.

\textbf{Event-anchored.} The model is given a video prefix that stops just before a collision and must select, from four candidates, the single frame showing the scene state \emph{right after that collision}. The hard distractors show the same scene at other moments, which prevents appearance-only matching and forces the model to predict the outcome of the collision. For example, after observing a gray and a yellow metal cylinder on a collision course over frames 10 to 30, the model must identify the state right after they collide at frame 40. The query names the anchor, \emph{``Which image is closest to the correct future state of the whole scene?''}, thereby pinning down the target moment rather than leaving any plausible later state correct. Construction variants tune difficulty and guard against shortcuts by altering the cut point, the source of the distractors, and the length of the prefix gap.

\textbf{Time-anchored.} The model is given a short observation and must identify the exact scene state at \emph{a precise later time}. This family probes temporal precision rather than event recognition, because no event needs to happen at that instant. For example, after watching a one-second observation over frames 0 to 25, the model must select the frame showing the scene 2 seconds later, at frame 78. The query is \emph{``Watch the 1-second observation video. Which image best shows what the scene will look like 2 seconds after the observation ends?''} The only variation is the prediction horizon, 1 second or 2 seconds.

\textbf{Window-anchored.} The model observes a brief window of activity and must identify the scene state \emph{a short horizon into the future}. Because every window item uses the same query, the items differ in the physical situation rather than the wording. The future window may contain no event, a single collision, an object entering or leaving view, or several events. For example, after observing frames 90 to 109, the model must select the frame at 125, by which point one collision has occurred. The query names the anchor, \emph{``Which image best matches the next scene state?''}, with the horizon fixed per item. Construction also varies the horizon and the distractor difficulty.

\section{The Retrieval Track}\label{app:retrieval}

The retrieval track (Figure~\ref{fig:retrieval}) uses the same
prefix and query as selection, but replaces the four candidates with the full
database, so a method cannot exploit the fact that exactly one of a small set is
correct. Distractors are not constructed per item here. They are whatever else the
database holds, which for a scene-dense database means many frames that differ
from the target only in the position of one or two objects.

The standard evaluation uses \nretqueries{} queries over \ndbimages{} frames.
Recall@$k$ and MRR are computed against the single correct frame, so chance
Recall@1 is 0.01\%, and we report the median rank of that frame because the mean
is dominated by a small number of very poorly ranked queries.

\paragraph{Query constructions.} The retrieval pool is generated by six builders that differ in what fixes the
target moment and where the observation is cut, as summarized in
Table~\ref{tab:retrieval_types}. Two builders are time-anchored, with the target set by a clock offset rather
than by an event.
Builder~1 asks for the final state of the clip and varies difficulty only through
the observation length, so a longer observation leaves less to predict.
Builder~2 fixes a two-second observation and sweeps the horizon. The remaining
four are event-anchored and differ in where the observation sits relative to the
collision. Builders~3 and~5 ask for the event frame itself, from an observation
that ends shortly before the event or a full second before it. Builder~4 cuts the
observation exactly at the collision and asks for the state one second later.
Builder~6 places the collision half a second after the observation ends, so the
event to be predicted is never seen. For query-specific hard negatives, every builder samples same-video wrong-time
frames, falls back to a denser offset grid when the first choices collide with
the target or with each other, and uses a fixed seed. Each builder caps its
source database at 30{,}000 frames, from which the \ndbimages{}-frame database of
the standard test set is drawn, so the pool sizes in
Table~\ref{tab:dynapix_stats} are sources rather than the evaluated database.
The retrieval builders use 25\,fps for time-to-frame conversion.

\begin{table}[t]
\centering\small
\setlength{\tabcolsep}{3pt}
\begin{tabular}{@{}cllc@{}}
\toprule
 & Observation & Target & Anchor \\
\midrule
1 & 0 to 2, 3 or 4\,s   & clip end (5\,s)      & time \\
2 & 0 to 2\,s           & obs end $+$1, 2, 3\,s & time \\
3 & 0 to 5 frames pre-event & the event frame  & event \\
4 & 1\,s, ends at event & event $+$1\,s        & event \\
5 & 1\,s, ends 1\,s pre-event & the event frame & event \\
6 & 1\,s, event 0.5\,s later & obs end $+$1\,s & event \\
\bottomrule
\end{tabular}
\caption{The six future-state retrieval builders. Time-anchored builders fix the
target with a clock offset, event-anchored builders fix it with a collision.
Builder~3 allows up to three events per video, builders~4 to~6 one.}
\label{tab:retrieval_types}
\end{table}

\section{Transfer Suite Construction}\label{app:transfer}

To measure transfer beyond our own simulator, we build zero-shot suites with the same oracle recipe on MoVi-A and Physion \citep{greff2022kubric,bear2021physion}. In these suites, simulator metadata or trial frames identify event times and target frames. MoVi-A is a Kubric rigid-body setting, and its multiple-choice suite contains 327 event-anchored and 1{,}198 time-anchored questions, for 1{,}525 in total.

The Physion suite is built from raw HDF5 trials and contains 583 event-anchored and 1{,}612 time-anchored questions, for 2{,}195 in total. Its event items come from three regimes: A1 contact, A2 motion and A3 landing. A1 targets first contact, A2 targets motion onset, and A3 targets landing on the ground. The prefix ends 10 frames before the event. Each item has four candidates: the correct frame, two same-video frames strictly after the correct one and one cross-video frame from the same scenario.

\section{Design Properties}\label{app:design}

\bench{} is designed as a physics-grounded and auditable benchmark for future-state prediction, with reduced reliance on surface appearance, and three construction choices support this goal. First, oracle frames provide exact labels through frame identity and timestamp. Second, same-scene wrong-time distractors reduce the usefulness of static matching in the time-anchored and window-anchored selection families. Third, multiple temporal regimes separate temporal precision from coarse event recognition. The benchmark also includes selection and retrieval tracks, counterfactual retrieval queries, and cross-simulator transfer suites. Synthetic scenes are a deliberate tradeoff for exact, controllable ground truth, as they allow prefix boundaries, event times, interventions and distractors to be specified with precision that is not available in ordinary web video.

\section{Implementation Details}\label{app:impl}

Table~\ref{tab:impl} lists the main implementation settings used for the reported
experiments. The reproduction-critical defaults not repeated here live in the
released configuration files.

\paragraph{Benchmark construction.} Every builder uses a fixed seed and an 80/20
train and test division of scenes, which yields the 9{,}368 and 2{,}208 counts of
Table~\ref{tab:dynapix_stats}. Time-anchored and window-anchored items take all
three distractors from the same video, while the event builder additionally offers
a cross-scene variant, the source of the cross-scene subtypes in
Table~\ref{tab:event_subtypes}. The event builder observes a 20-frame clip by
default and stops 10 frames before the collision, requires at least 10 frames
between the answer frame and the end of the clip, and draws same-video distractors
at 16, 32 and 48 frames on either side of the target. The window builder observes
16 frames and looks 8 frames ahead by default. Observation length is configurable
per subtype, so the worked examples in Appendices~\ref{app:subtypes}
and~\ref{app:families} show items whose spans differ from these defaults. The
selection time builder treats 26 frames as one second, matching CLEVRER's 128
frames over roughly five seconds, while the retrieval builders convert at
25\,fps, a difference that moves a one-second target by at most one frame.

\paragraph{Rationale construction.} The oracle-to-predicate stage calls an object
stationary when its speed falls below 0.05 in simulator units, cuts the prefix 12
frames before the first collision, discards scenes whose prefix would be shorter
than 5 frames, and inspects the 5 frames before the cut to decide which pairs are
on a collision course. Time-to-event is bucketed rather than stated numerically,
as imminent within 15 frames and soon within 40. The teacher is served locally
and sampled at temperature 0.7, and a rationale is rejected and resampled unless
it falls between 80 and 250 words. This filter is intended to reduce answer
restatement and unsupported detail.

\paragraph{Training.} Students are trained with LoRA adapters over a 4-bit
NF4-quantized base, with the vision encoder frozen and adapters on the attention
and feed-forward projections of the language tower. The contrastive retriever
finetune keeps the same training infrastructure and replaces the generative
objective with a contrastive ranking one. It uses a cached multiple-negatives
ranking objective wrapped in a Matryoshka loss over seven nested embedding widths
from 4{,}096 down to 64, so one finetune yields a family of truncatable
embeddings. The cached formulation permits more negatives than the micro-batch
size alone would provide.

\begin{table}[t]
\centering\small
\setlength{\tabcolsep}{4pt}
\resizebox{\columnwidth}{!}{%
\begin{tabular}{@{}ll@{}}
\toprule
Setting & Value \\
\midrule
\multicolumn{2}{@{}l}{\emph{Student training}} \\
\quad Base models & Qwen3-VL 8B, 2B \\
\quad LoRA rank / alpha & 16 / 32 \\
\quad Base quantization & 4-bit NF4 \\
\quad Learning rate & $2\times10^{-5}$ \\
\quad Epochs & 3 \\
\quad Micro batch / accumulation & 1 / 16 \\
\quad Max sequence length & 3{,}072 \\
\quad Precision & bf16 \\
\midrule
\multicolumn{2}{@{}l}{\emph{Visual input}} \\
\quad Frames per prefix & 12 \\
\quad Frame resolution & $336\times336$ \\
\midrule
\multicolumn{2}{@{}l}{\emph{Teacher}} \\
\quad Model & Qwen3-32B \\
\quad Temperature & 0.7 \\
\quad Rationale length & 80 to 250 words \\
\midrule
\multicolumn{2}{@{}l}{\emph{Contrastive retriever}} \\
\quad Base model & Qwen3-VL-Emb-8B \\
\quad Loss & Matryoshka over cached MNRL \\
\quad Embedding widths & 4{,}096 to 64, seven levels \\
\quad Negatives & in-batch and same-scene \\
\midrule
\multicolumn{2}{@{}l}{\emph{Inference}} \\
\quad Instruct decoding & greedy \\
\quad Thinking decoding & $T{=}0.6$, top-$p$ 0.95, top-$k$ 20 \\
\quad Max new tokens & 512 instruct, 8{,}192 thinking \\
\bottomrule
\end{tabular}}
\caption{Main implementation settings for the reported experiments. Frames per
prefix is the number of frames sampled from the observation as model input, not
the length of the observation itself.}
\label{tab:impl}
\end{table}

\section{Evaluation Protocol}\label{app:protocol}

Generative models receive the prefix as a video, the query as text, and the four
candidates as four separately labelled images in a randomized order, and are
asked to return a structured record rather than a bare letter. The requested
record contains the forced choice, a self-reported confidence, a distribution
over the four options, a second choice, an ambiguity tag, one short rationale per
option, and a final justification. We score only the forced choice. The remaining fields
exist so that a refusal, a hedge, or a tie is visible in the output rather than
hidden inside a scalar, and so that an unparseable answer can be distinguished
from a wrong one.

We nonetheless count unparseable outputs as wrong. A model that emits no
extractable choice has not identified a future state, and discarding those cases
would flatter exactly the models that fail to commit. This matters for one row of
Table~\ref{tab:selection}, where the thinking base returns a parseable record for
only 66.4\% of items, so its reported accuracy mixes genuine errors with
non-answers. We report the parse rate alongside the accuracy so the two can be
separated. Thinking models are allowed 8{,}192 new tokens against 512 for instruct models, a
budget sixteen times larger, which reduces but does not by itself eliminate
truncation as an explanation for a missing answer. Similarity scorers bypass this
protocol because they emit a score for each candidate and rank by construction.

\section{Human Study Details}\label{app:human}

Section~\ref{ssec:human} reports the three-annotator majority vote used as a
validity probe for the time-anchored family. Here we give the per-annotator
breakdown, agreement, and timing behind that summary. The three annotators
collectively judged a balanced subsample of 59 items drawn across CLEVRER, MoVi-A,
and Physion, seeing the prefix, query, and four candidates in randomized order with
the correct answer balanced across positions, for 127 judgments in total. No annotator saw any answer
key, and the interface exposed only the media the released form serves.

Table~\ref{tab:human_detail} gives per-annotator accuracy by family. Every
annotator is far above the 25\% chance rate on time-anchored items and far above
the best zero-shot model there (83.3\% against 27.3\%), which is the comparison the
validity probe rests on. On event-anchored items annotators are at or below the
models, at 58.3\%, 40.0\% and 63.6\%, and they are also slowest there, with a
median decision time of 39.9\,s against 15.8\,s on time. Agreement over the items
all three saw is fair (Fleiss' $\kappa=0.37$, 19 overlapping items). Annotation
depth is uneven: of the 59 items, 10 were seen by one annotator, 30 by two and 19
by all three, so the majority vote reduces to a single judgment on the first group
and can be a tie on the second. We therefore read the per-annotator columns as the
primary human evidence and the vote as a summary.

\begin{table}[t]
\centering\small
\setlength{\tabcolsep}{7pt}
\begin{tabular}{@{}lrrr@{}}
\toprule
Annotator & Event & Time & Window \\
\midrule
A1 & 58.3 & 83.3 & 60.9 \\
A2 & 40.0 & 100.0 & 81.8 \\
A3 & 63.6 & 83.3 & 81.2 \\
\midrule
Majority vote & 58.3 & 83.3 & 73.9 \\
\bottomrule
\end{tabular}
\caption{Per-annotator human accuracy (\%) by family, with the majority vote used
in Table~\ref{tab:selection}. Median decision times are 39.9\,s (event), 15.8\,s
(time), and 20.3\,s (window).}
\label{tab:human_detail}
\end{table}

\section{What the Rationale Contributes}\label{app:ablation}

\begin{table}[t]
\centering\small
\setlength{\tabcolsep}{6pt}
\begin{tabular}{@{}lrrr@{}}
\toprule
 & CLEVRER & MoVi-A & Physion \\
\midrule
Full student & 87.5 & 39.9 & 64.6 \\
\quad $-$ candidate analysis & 88.0 & 35.9 & 69.0 \\
\quad $-$ match layout & 86.8 & 40.3 & 68.9 \\
\bottomrule
\end{tabular}
\caption{Component ablation of the distilled student, overall accuracy (\%) in
reasoning mode. The student writes a rationale before answering. Each row removes
one component from the full model.}
\label{tab:ablation}
\end{table}

Two packaging choices could explain the gain instead of grounding, so we remove
each. One drops the explicit candidate-analysis step from the rationale and is
retrained, and the other removes the match-style answer layout and is tested with
the baseline layout. Table~\ref{tab:ablation} reports overall accuracy in domain
and on the two transfer suites, with the student in reasoning mode. These transfer
figures are not comparable to the answer-mode numbers above. In domain, neither
component is worth one point, so the in-domain gain comes from the grounded
rationale rather than either formatting choice. On transfer, effects are small and
mixed: removing candidate analysis lowers MoVi-A but raises Physion, so neither
component is uniformly responsible for transfer. Together with the answer-only row
of Table~\ref{tab:distill}, this places the gain in the content of the
supervision rather than in its presentation.

\section{Qualitative Examples of Grounded Rationales}\label{app:rationales}

Section~\ref{ssec:method-pipeline} describes how a rationale is built from the
simulator's record rather than sampled from a teacher's guess. Because that
distinction is the core of the method, we give a complete rationale here so it can
be inspected directly. The example is an event-anchored item from CLEVRER video
86, where the prefix is cut before a collision and the target is frame 63. Every
statement below is a narration of oracle predicates, and the validator has removed
all raw coordinates, timestamps, and numeric values, so the student is supervised
only in terms it can recover from pixels at test time.

\vspace{2pt}
\noindent\textbf{Query.} \emph{``Which image is closest to the correct future state
of the whole scene?''}

\vspace{2pt}
\noindent\textbf{Scene parse.} At the end of the observation, the yellow rubber
sphere is near the right side of the scene, moving slowly toward the southwest.
The blue metal cube is near the top-right, moving quickly toward the south. The
yellow metal cylinder is near the top and stationary, and the gray metal sphere is
also near the top. The yellow rubber sphere and the blue metal cube are on a
collision course.

\vspace{2pt}
\noindent\textbf{Motion analysis.} The yellow rubber sphere is moving slowly toward
the southwest, so given its speed and direction it will shift only slightly
downward and to the left, and because the collision is imminent near the top-right
it will not have moved far by the target moment. The stationary objects remain
where they are unless struck.

\vspace{2pt}
\noindent\textbf{Prediction.} At the collision moment, the yellow rubber sphere and
the blue metal cube will be very close near the top-right, while the yellow metal
cylinder remains near the top, the gray metal sphere remains near the top, and the
brown metal cube remains near the top-left.

\vspace{2pt}
\noindent\textbf{Outcome.} The yellow rubber sphere and the blue metal cube are
very close near the top-right, while the yellow metal cylinder is near the top and
the gray metal sphere is near the top.

\vspace{2pt}
Two properties of this rationale are worth noting. First, it is stated entirely in
qualitative terms, such as \emph{near the top-right} and \emph{moving slowly toward
the southwest}, so nothing in it could be recited from a coordinate the student
cannot observe. Second, the candidate it selects is fixed by the simulator's record
rather than by plausibility, so a fluent but physically wrong narration cannot be
produced by construction. This is what separates the supervision from ordinary
chain-of-thought distillation, in which the rationale records only what the teacher
guessed.

\end{document}